%% file: main.tex
\pdfoutput=1
\documentclass{article}
\usepackage{iclr2021_conference,times}
\PassOptionsToPackage{numbers}{natbib}

\usepackage[utf8]{inputenc} 
\usepackage[T1]{fontenc}    
\usepackage{hyperref}       
\usepackage{url}            
\usepackage{booktabs}       
\usepackage{amsfonts}       
\usepackage{nicefrac}       
\usepackage{microtype}      
\usepackage{comment}        
\usepackage{amsmath}
\usepackage{multirow}
\usepackage{bm}
\usepackage{amsthm}
\usepackage{hhline}
\usepackage{amssymb}
\usepackage{makecell}
\usepackage{mathtools}
\usepackage{color}
\usepackage{xcolor}
\usepackage{MnSymbol}
\usepackage{makecell}
\usepackage{graphicx}
\graphicspath{ {./images/} }
\usepackage{arydshln}
\usepackage{booktabs}
\usepackage{framed}
\usepackage[font=small]{caption}
\usepackage[scientific-notation=true]{siunitx}
\usepackage{wrapfig}
\usepackage{algorithm}
\usepackage{algorithmic}
\usepackage{lipsum}
\usepackage{sidecap}
\usepackage{pifont}

\newcommand{\namel}{\textsc{Ask \& Explore}}
\newcommand{\names}{\textsc{AnE}}

\definecolor{gg}{RGB}{15,150,15}
\definecolor{rr}{RGB}{230,45,45}

\makeatletter
\def\maketag@@@#1{\hbox{\m@th\normalfont\normalsize#1}}
\makeatother

\input{math_commands.tex}

\AtBeginDocument{%
  \addtolength\abovedisplayskip{-0.3\baselineskip}%
  \addtolength\belowdisplayskip{-0.3\baselineskip}%
}

\title{Ask \& Explore: Grounded Question Answering for Curiosity-Driven Exploration}

\author{%
  Jivat Neet Kaur$^\heartsuit$, Yiding Jiang$^\spadesuit$, Paul Pu Liang$^\spadesuit$\\
  $^\heartsuit$Birla Institute of Technology and Science, Pilani\\
  $^\spadesuit$Carnegie Mellon University\\
  \texttt{}\\
}

\iclrfinalcopy 
\begin{document}

\maketitle

\begin{abstract}
In many real-world scenarios where extrinsic rewards to the agent are extremely sparse, curiosity has emerged as a useful concept providing intrinsic rewards that enable the agent to explore its environment and acquire information to achieve its goals. Despite their strong performance on many sparse-reward tasks, existing curiosity approaches rely on an overly holistic view of state transitions, and do not allow for a structured understanding of specific aspects of the environment. In this paper, we formulate curiosity based on grounded question answering by encouraging the agent to ask questions about the environment and be curious when the answers to these questions change. We show that natural language questions encourage the agent to uncover specific knowledge about their environment such as the physical properties of objects as well as their spatial relationships with other objects, which serve as valuable curiosity rewards to solve sparse-reward tasks more efficiently.
\end{abstract}

\vspace{-4mm}
\section{Introduction}
\vspace{-3mm}

Efficient exploration in the absence of dense reward signals is a long-standing problem in reinforcement learning~\citep{vecerik2017leveraging}. Without dense extrinsic signals, a promising alternative is to define suitable auxiliary \textit{intrinsic} signals that can help the agent in exploring its environment~\citep{laud2004theory}. Recently, \textit{curiosity} has emerged as a promising computational framework for modeling intrinsic reward and has brought major advances in many sparse-reward domains~\citep{pathak2017curiosity,burda2018large,burda2018exploration,pathak2019self,dean2020see}. While the algorithmic details of different methods vary, the core idea is to use changes in the observed state as the intrinsic reward to encourage agents to explore their environment. Despite their strong performance on many sparse-reward tasks, these existing approaches tend to rely on a holistic view of state transitions and do not allow for a targeted understanding of specific aspects of the environment. However, not all states are equally interesting but such information is not available to the agent a priori. On the contrary, humans rely on extensive knowledge about the world when exploring the environment. Language serves as a powerful medium for encoding this knowledge. A particular type of language that humans use is \textit{question} -- in an unfamiliar environment, humans often start the exploration by asking what can be done in the environment.
Based on this observation, we hypothesize that \textit{language-based question answering} may provide a grounded and targeted medium to probe specific knowledge about the current state in order to solve the task at hand.

As a step towards more structured and flexible curiosity-driven learning, we develop a novel form of curiosity, \namel\ (\names), that leverages \textit{grounded question answering} to encourage the agent to ask questions about the environment and be curious when the answers to these questions change. These questions can capture physical properties of the objects (e.g., \textit{Is the large sphere green in color?}) as well as their spatial relationships with other objects (e.g., \textit{Are there any blue spheres behind the cyan ball?}). By using language as a compositional medium to uncover specific knowledge about the environment, we are able to train an agent that explores and solves challenging long-horizon sparse-reward tasks. In addition to our qualitative results, we perform an in-depth study of what type of questions are useful under what scenarios to provide empirical guidelines for applying our method.

\vspace{-3mm}
\section{Background}
\vspace{-3mm}

\textbf{Exploration bonuses and curiosity-driven exploration}. Exploration bonuses motivate agents to explore their environment even when extrinsic reward $r_t^e$ is sparse (or zero) by training the policy to maximize a new reward $r_t = r_t^e + r_t^i$, where $r_t^i$ is the exploration bonus or the intrinsic reward at time $t$~\citep{krebs2009novelty,dayan1996exploration,sutton1990integrated}. The intrinsic reward $r_t^i$ is designed to be higher in novel states in order to encourage the agents to explore less frequently visited states. In recent years, several promising algorithms in this family include: 1) Curiosity-driven exploration by self-supervised prediction~\citep{pathak2017curiosity,burda2018large,pathak2019self}, which formulates an intrinsic reward that encourages the agent to favor transitions with high prediction error using dynamics-based learning, and 2) Random Network Distillation (RND)~\citep{burda2018exploration}, which encourages novelty by training the policy to minimize the prediction error of a predictor neural network as it tries to mimic a randomly initialized target neural network. Several other approaches have used count-based exploration~\citep{bellemare2016unifying, tang2017exploration} and multimodal signals~\citep{dean2020see} to encourage exploration.

\textbf{CLEVR-Robot Environment:} We perform experiments using the CLEVR-Robot Environment~\citep{jiang2019language}, an open-source object interaction environment built using the MuJoCo physics engine~\citep{todorov2012mujoco} and CLEVR language engine~\citep{johnson2017clevr}. The environment is designed to serve as a testbed for studying grounded language understanding and object manipulation. To succeed in this environment, the agent must be able to handle a varying number of objects with diverse visual and physical properties (see details in Appendix~\ref{app_dataset}).

\begin{figure}[tbp]
\centering
\vspace{-6mm}
\includegraphics[width=\linewidth]{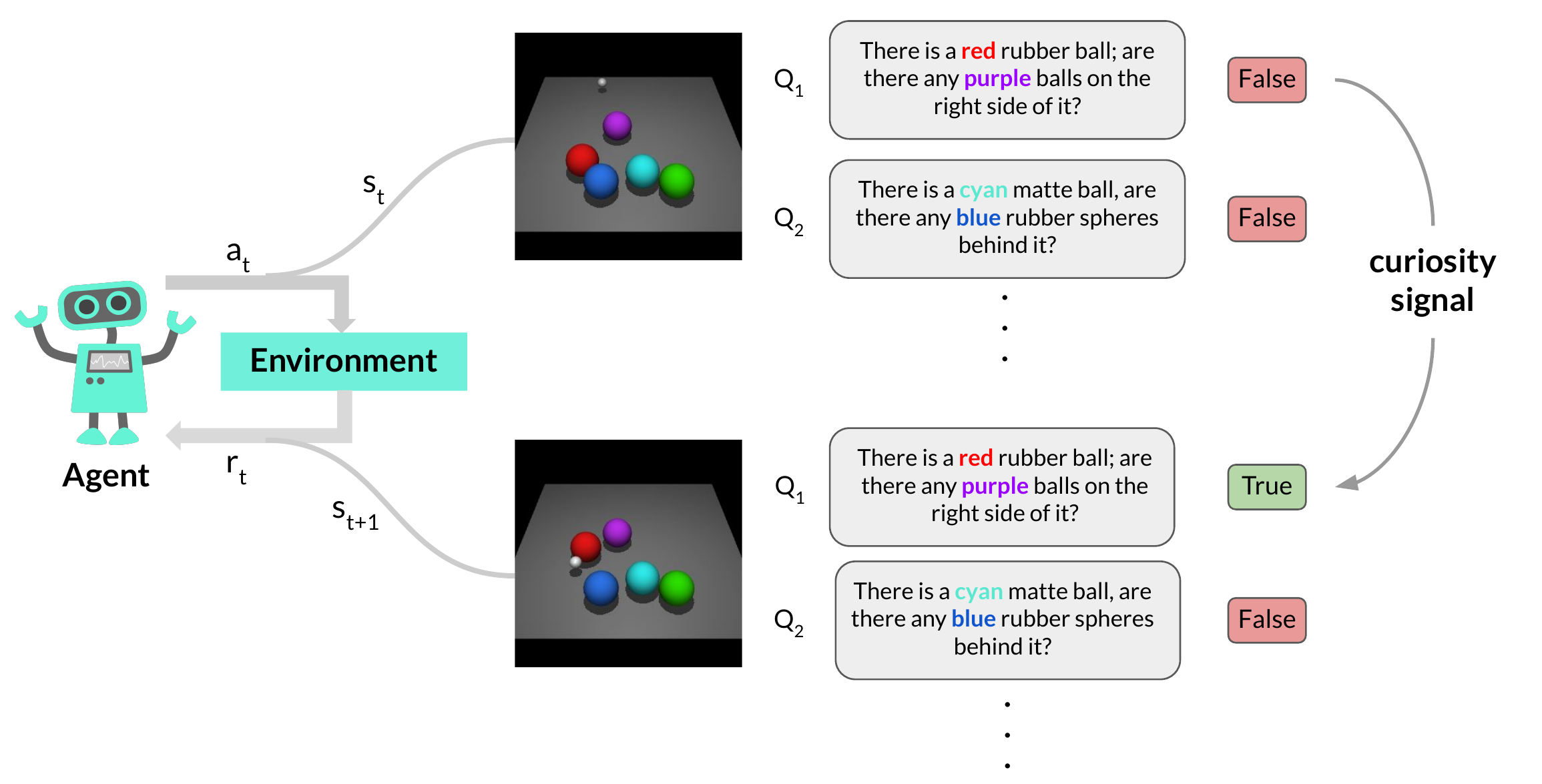}
\caption{\textbf{\namel}: Our approach proposes a curiosity formulation that leverages grounded question answering to query specific knowledge about the environment. The agent is encouraged to ask questions and be curious about transitions when the answer to a question changes (details in Section~\ref{sec: approach}).\vspace{-6mm}}
\label{fig:approach}
\end{figure}

\vspace{-3mm}
\section{\namel}
\label{sec: approach}
\vspace{-3mm}

Our goal is to develop an intrinsic reward that leverages the knowledge about the physical properties of objects and how objects in the environment relate to each other. Such intrinsic reward may bridge the gap between passive pattern recognition and active decision making. To design the intrinsic reward, we chose grounded language as a flexible medium for encoding this knowledge. The CLEVR-Robot environment provides an ideal testbed for using grounded language as a source of intrinsic reward as it provides functionalities to generate scenes and language (in the form of questions) that can be evaluated as the agent interacts with the environment. Note that access to the true state and language is not required. Indeed, for an agent in the real world, such assumptions do not hold. Nonetheless, just like humans can describe a scene with language, the agent can also be equipped with a parameterized visuolinguistic model such as a visual question answering model (VQA) or image captioning model. We plan to explore these directions in future works.


We formulate an intrinsic reward that aims to generate the agent's curiosity about transitions when the answers to the agent's questions grounded in the environment change. At every step, the agent has access to $n$ questions, $q_1, q_2, ... q_n$. For question $q_k$, the difference in the answer before the transition $A(s_t, q_k)$, and after the transition $A(s_{t+1}, q_k)$ contributes to the curiosity signal corresponding to that action (Figure \ref{fig:approach}). To evaluate $A(s_t, q_k)$, we experiment with two types of intrinsic rewards that leverage language. One of them uses the labeling function of the CLEVR-Robot environment and the other utilizes a parameterized VQA model (experiments with the latter can be found in Appendix \ref{app_vqa}). The intrinsic reward at time $t$, $r^i_t$, expressed as
\begin{equation}
     r_t^i = \sum_{k=1}^n \mathbbm{1}[A(s_t, q_k) \neq A(s_{t+1}, q_k)]
\end{equation}

Further algorithmic details such as how the questions are selected can be found in Appendix \ref{imp_details}.


\begin{figure}[tbp]
\centering
\begin{subfigure}[t]{0.24\columnwidth}
\centering
\includegraphics[width=0.99\linewidth, height=0.8\linewidth]{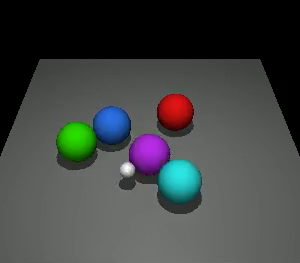}
\vspace{-0.5cm}
\caption{\scriptsize \centering Goal is: ``{\it There is a \textbf{\textcolor{green}{green}} sphere; are there any rubber \textbf{\textcolor{cyan}{cyan}} balls \textbf{in front of} it?}".}
\end{subfigure}
\hfill
\begin{subfigure}[t]{0.24\columnwidth}
\centering
\includegraphics[width=0.99\linewidth, height=0.8\linewidth]{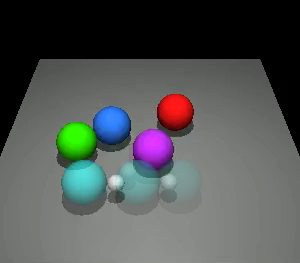}
\vspace{-0.5cm}
\caption{\scriptsize \centering Agent performs actions and interacts with the environment and tries to satisfy goal.}
\end{subfigure}
\hfill
\begin{subfigure}[t]{0.24\columnwidth}
\centering
\includegraphics[width=0.99\linewidth, height=0.8\linewidth]{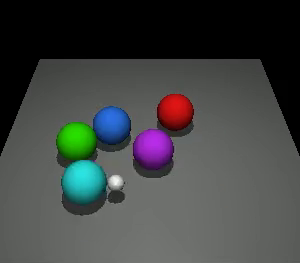}
\vspace{-0.5cm}
\caption{\scriptsize \centering Resulting state: Agent receives \textbf{+1} reward.}
\end{subfigure}
\hfill
\begin{subfigure}[t]{0.24\columnwidth}
\centering
\includegraphics[width=0.99\linewidth, height=0.8\linewidth]{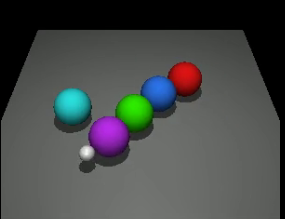}
\vspace{-0.5cm}
\caption{\scriptsize \centering Object ordering}
\end{subfigure}
\caption{We use the CLEVR-Robot environment and consider both dense (a-c) and sparse (d) reward settings. The global location of the objects vary across episodes (images from~\citet{jiang2019language}).\vspace{-4mm}}
\label{fig:clevr_env}
\end{figure}

\vspace{-3mm}
\section{Experiments}
\vspace{-3mm}

We design our experiments to understand the following overarching question: \textit{Does an agent with grounded language understanding explore the environment in a more structured and efficient manner?} To answer different aspects of this question, we first evaluate our approach in tasks with different reward sparsity (Section \ref{sec: varying_sparsity}). Then, we evaluate the impact of different types of grounded language understanding on the performance and how the impact differs in settings with varying reward sparsity (Section \ref{sec: varying_complexity}). Finally, we study the effect of the linguistic feedback's density on the efficacy of exploration (due to space limit, we defer the details to Appendix \ref{app_multiple_ques}).

We compare our approach to three baselines:

1. Proximal Policy Optimization (PPO)~\citep{schulman2017proximal} (no exploration bonus)

2. Intrinsic Curiosity Module (ICM)~\citep{pathak2017curiosity}

3. Random Network Distillation (RND)~\citep{burda2018exploration}.

We use the same optimized hyperparameters from the original papers~\citep{pathak2017curiosity, burda2018exploration}. The agent is trained using PPO in all experiments with the same hyperparameters\footnote{The original ICM uses A3C but we used PPO similar to~\cite{burda2018large}.}. We perform three independent runs of each algorithm without any tuning of random seeds, and plot the mean and standard deviation across the three runs (see details in Appendix~\ref{app_details}).


\vspace{-2mm}
\subsection{Varying Degree of Reward Sparsity}
\label{sec: varying_sparsity}
\vspace{-2mm}

To study in what scenarios grounded language understanding can help exploration, we test our approach in two tasks with drastically different reward sparsity.

\textbf{Dense reward setting.} \quad In this setting, the agent needs to complete an object alignment goal where the spatial relationship between two objects in the environment is specified, for example, \textit{"There is a green sphere; are there any rubber cyan balls in front of it?"} (Figure \ref{fig:clevr_env} (a-c)). Goal descriptions take the form of questions which can be evaluated on the state to assess if the goal has been met successfully. The agent receives a reward of $+1$ if it manipulates the objects to achieve the desired spatial arrangement. When the environment is reset before every episode, it is ensured that the goal state is not satisfied initially. 

\textbf{Sparse reward setting.} \quad In the \textbf{object ordering} task (Figure \ref{fig:clevr_env} (d)), the agent needs to order the objects by color in a single line, for example, \textit{“arrange the objects so that their
colors range from blue to green in the horizontal direction, and keep the objects close vertically"}. The ordering of colors we specify is: \textit{cyan, purple, green, blue, red} from left to right. The agent is given a $+10$ reward if it is able to successfully order the objects in this arrangement, and $0$ otherwise. The rationale behind opting for a $+10$ reward instead of $+1$ is to compensate for the extreme reward sparsity which led to the agent not making any significant progress.

\begin{figure}[tbp]
\begin{minipage}{.49\textwidth}
\centering
\vspace{-2mm}
\includegraphics[width=\linewidth]{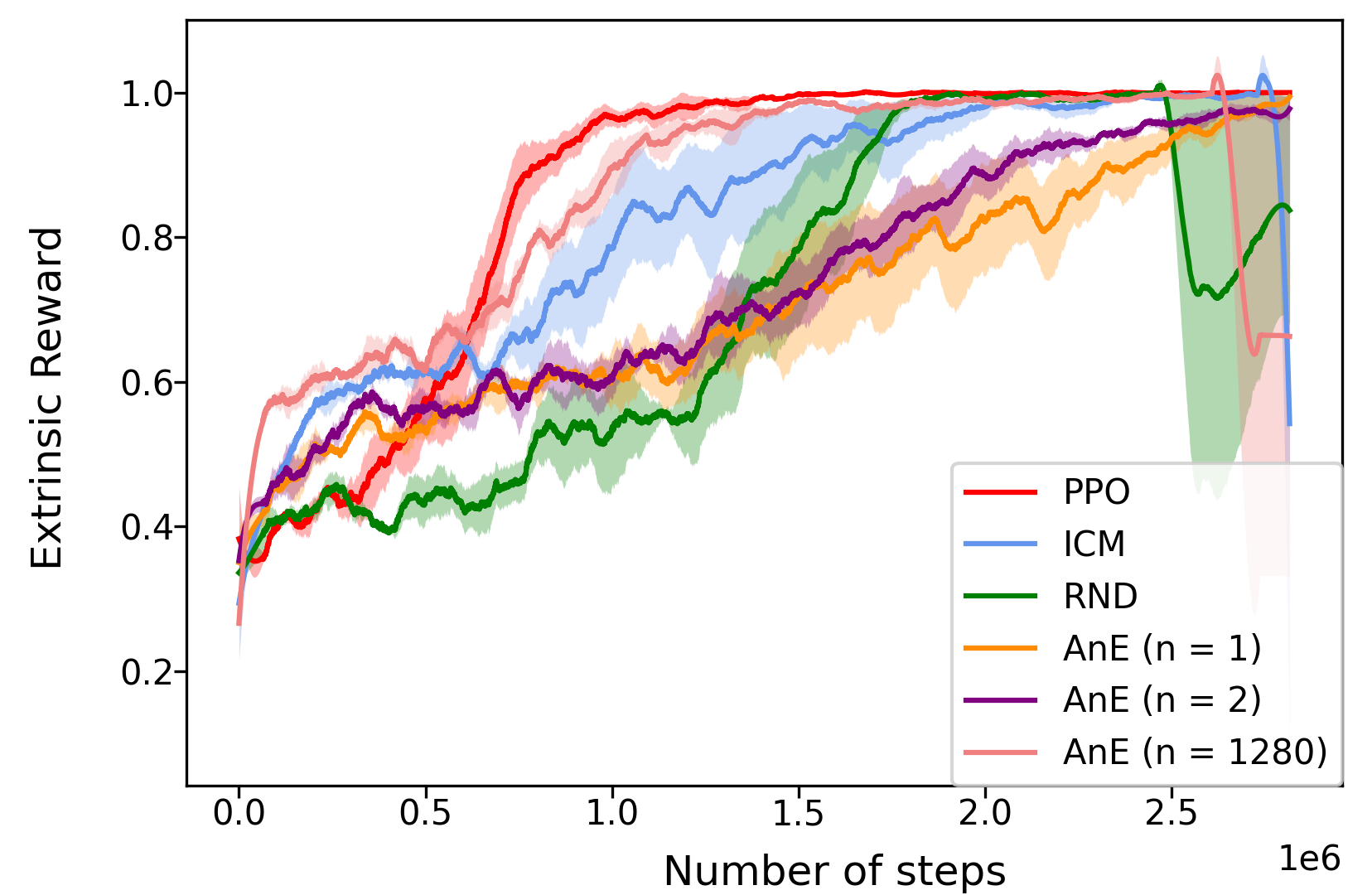}
(a) Dense reward setting
\end{minipage}%
\hspace{.02\textwidth}
\begin{minipage}{.49\textwidth}
\centering
\vspace{-2mm}
\includegraphics[width=\linewidth]{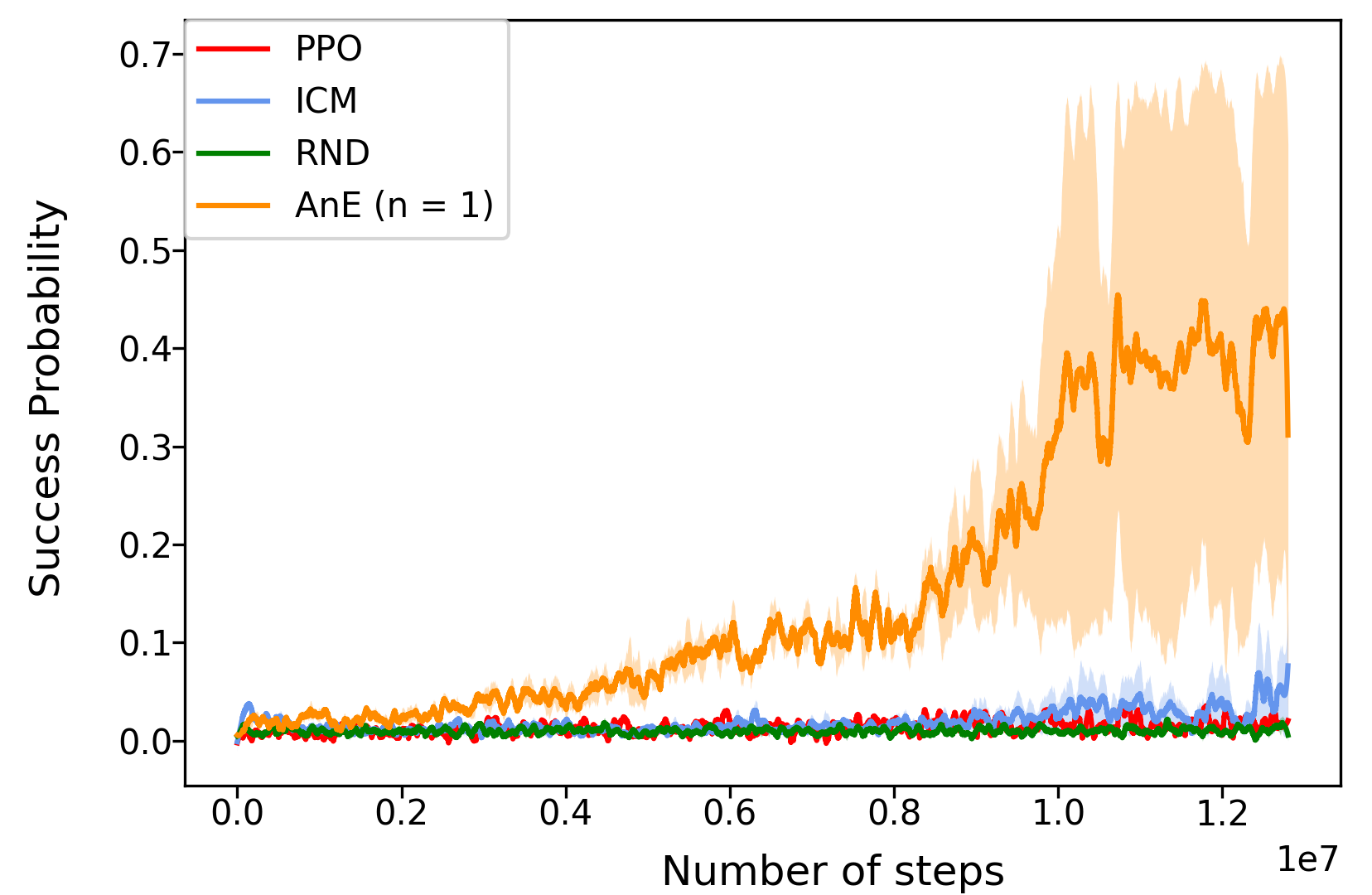}
(b) Sparse reward setting
\end{minipage}%
\caption{\names\ significantly outperforms the baselines (PPO, ICM and RND) in the sparse reward setting which demonstrates the effectiveness of an intrinsic reward based on grounded question answering.\vspace{-6mm}}
\label{fig:sparsity_plots}
\end{figure}

The results for the two settings are shown in Figure \ref{fig:sparsity_plots}. We observe in the dense reward setting, PPO with no exploration bonus outperforms all curiosity-driven methods, which highlights that curiosity does not provide a significant advantage when the reward is dense. In the sparse reward task, we find that while all existing baseline methods struggle to make meaningful progress, \names\ significantly outperforms the baselines in this setting using a single question ($n = 1$) at each step. This confirms our hypothesis -- an intrinsic reward that leverages grounded language understanding is better at exploring the environment. The exploration results in a wider coverage of relevant states and helps the agent learn to solve the task more efficiently
compared to existing novelty-based exploration methods.
In addition, we study the impact of scaling to multiple questions in both dense and sparse reward environments in Appendix \ref{app_multiple_ques}, and the performance of different approaches in the absence of any extrinsic reward in Appendix \ref{app_intrinsic_only}.

\vspace{-2mm}
\subsection{Varying complexity of questions}
\label{sec: varying_complexity}
\vspace{-2mm}

To better understand which questions are most useful for the task, we test the performance of \names\ using three types of questions querying varying complexity of spatial relationships between objects in the environment - \textbf{one}, \textbf{two} and \textbf{three} ``hop'' questions:

\vspace{-1mm}
\textbf{One-hop}: \textit{``There is a red metallic sphere; are there any green matte balls left of it?''}

\vspace{-1mm}
\textbf{Two-hop}: \textit{``Are there any purple rubber balls that are on the left side of the red sphere that is behind the blue matte ball?''}

\vspace{-1mm}
\textbf{Three-hop}: \textit{``There is a green rubber ball behind the red metallic sphere; are there any blue balls in front of the purple matte sphere?''}

\begin{figure}[!ht]
\begin{minipage}{.49\textwidth}
\centering
\vspace{-2mm}
\includegraphics[width=\linewidth]{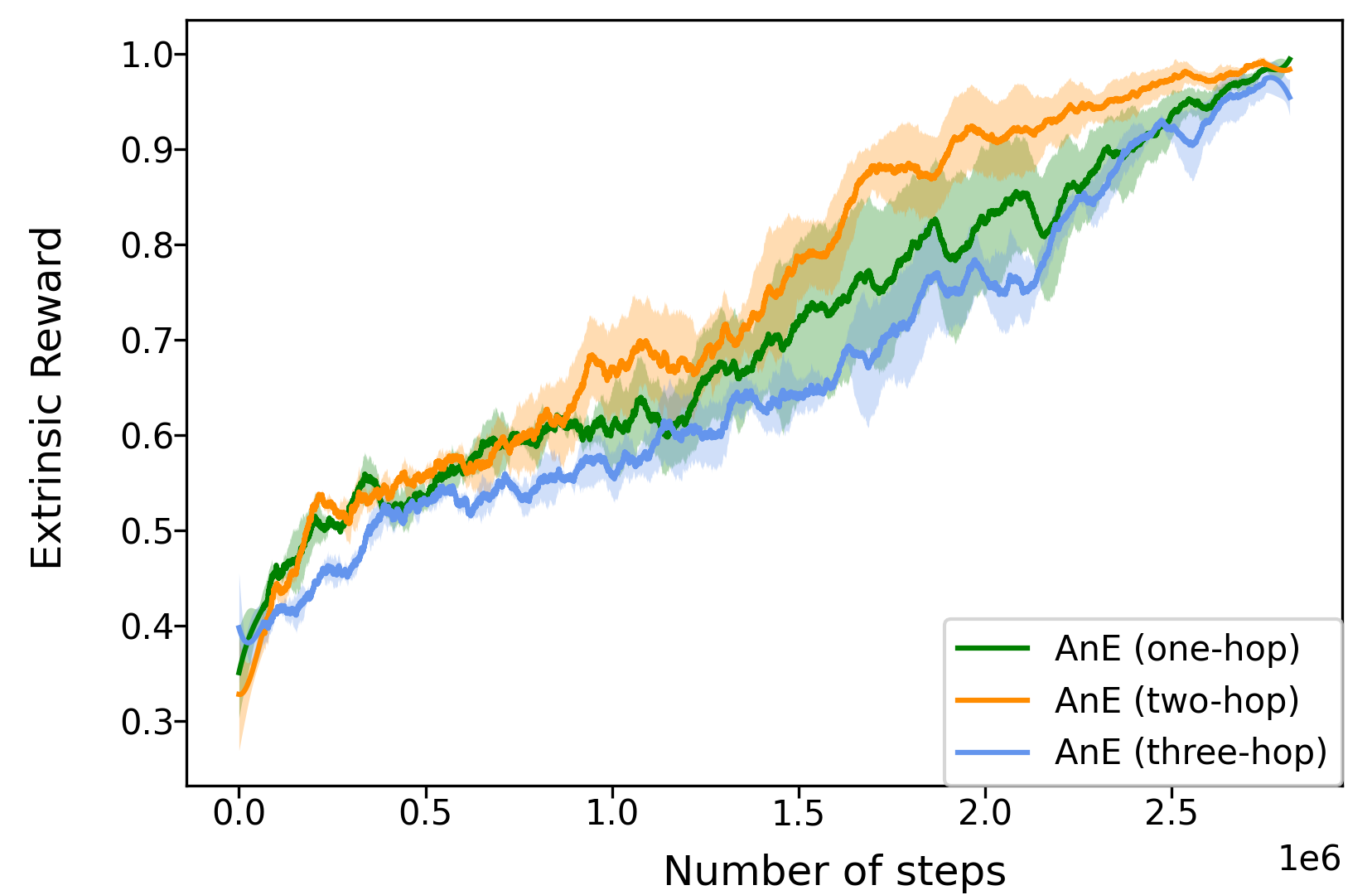}
(a) Dense reward setting
\end{minipage}%
\hspace{.02\textwidth}
\begin{minipage}{.49\textwidth}
\centering
\vspace{-2mm}
\includegraphics[width=\linewidth]{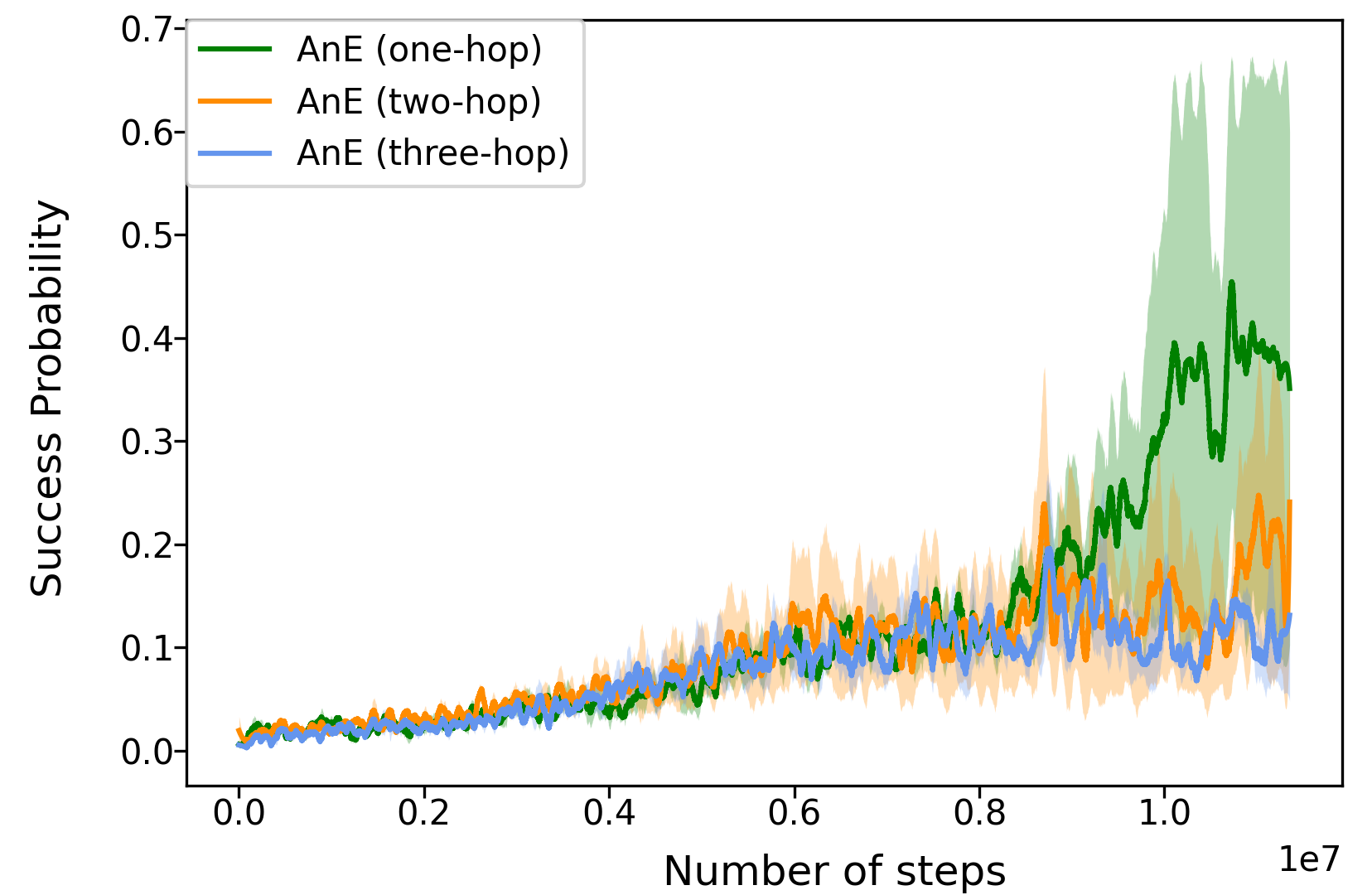}
(b) Sparse reward setting
\end{minipage}%
\caption{Comparing the performance of \names\ for questions of varying complexity. \vspace{-4mm}}
\label{fig:complexity_plots}
\end{figure}

We observe in Figure \ref{fig:complexity_plots} how the performance of the agent varies with increasing complexity of questions (in terms of a greater number of pair-wise relationships between objects). It is interesting to note that the relationship between language and agent performance is not the same across different task settings. The extrinsic reward increases in the dense reward setting as we move from one-hop to two-hop questions in the environment and then decreases as we progress to three-hop questions. While in the sparse task, the success rate is highest using one-hop questions and decreases as we increase the number of object relationships, although all improve over the baselines. Therefore, even simple probes of spatial relationships are sufficient as a curiosity signal.









\vspace{-3mm}
\section{Conclusion}
\vspace{-3mm}

In this paper, we proposed a novel form of curiosity by encouraging the agent to ask questions about the environment and be curious when the answers to their questions change. We show that this formulation of intrinsic reward probes targeted knowledge about the physical properties of the objects as well as their spatial relationships with other objects, achieving significantly better performance than existing curiosity methods on highly sparse reward tasks.



{\small
\bibliography{main}
\bibliographystyle{iclr2021_conference}
}

\clearpage
\onecolumn
\appendix

\section*{Appendix}

\vspace{-2mm}
\section{CLEVR-Robot Environment}
\label{app_dataset}
\vspace{-2mm}

The CLEVR-Robot environment~\citep{jiang2019language} was designed in MuJoCo for object manipulation tasks. In the environment, the agent can interact with objects with diverse visual and physical properties. The environment supports CLEVR style~ \citep{johnson2017clevr} language so the agent may receive linguistic feedback as it interacts with the environment.

We use the \textit{discrete} action space which consists of a point mass agent pushing 1 of 5 objects in 1 of the 8 cardinal directions for a fixed number of frames, so the discrete action space has size 40. Please refer to \citep{jiang2019language} for details of the action space.
As a proof of concept for using grounded language for exploration, we consider the standard 5 objects setting which contains a fixed set of 5 spheres of different colors- \textit{cyan, purple, green, blue, red.} We plan to experiment with the diverse objects setting where objects can take up different shapes such as \textit{cube, sphere} and \textit{cylinder} in future work.

The environment supports all CLEVR style language. For the experiments, we consider 3 types of language statement: \textit{one-hop}, \textit{two-hop} and \textit{three-hop}. The number indicates the number of objects involved in the spatial reasoning (for an \textit{h-hop question}, \textit{h+1} objects are involved) and, indirectly, the complexity of reasoning. The number of hops also affects the density of intrinsic reward and what kind of states the agent is encouraged to visit.

\vspace{-2mm}
\section{Implementation Details}
\label{imp_details}
\vspace{-2mm}

We begin with a fixed set of questions $S$ from which a subset of $n$ questions, $Q_1, Q_2, ... Q_n$ is sampled at each step of the episode. A counter is maintained for a large reservoir of possible questions to record the frequency of answer flips corresponding to each question. We use a hyperparameter $0.5 \leq \alpha \leq 1$ as a threshold to set an upper bound on the \% of answer flips any question can encounter when it is sampled, after which it is replaced by a new question sampled from the reservoir (e.g., If $\alpha = 0.6$ and $Q_1$ has witnessed 650 answer flips out of the 1000 times it was sampled, it is replaced by a new question $Q_k$ which has not been seen by the agent yet). This is an attempt to ensure that if the agent has learned transitions to exploit a particular language statement, it does not continue to exploit it.

Algorithm \ref{alg:ane} provides a more complete picture of the approach.

\begin{algorithm}[!ht]
    \caption{\names\ pseudo-code}
    \label{alg:ane}
    \begin{algorithmic}
        \STATE $N \gets$  number of rollouts
        \STATE $N_{\text{opt}} \gets$ number of optimization steps
        \STATE $K \gets$ length of rollout
        \STATE $S \gets$ set of questions initialized 
        \STATE $M \gets$ number of questions initialized in \textit{S}
        \STATE $C \gets$ counter for questions in \textit{S} initialized to 0 
        \STATE $n \gets$ number of questions queried at each step
        \STATE $D \gets$ initialized with $n*K$ questions from \textit{S}
        \STATE $\alpha \gets$ threshold which determines maximum answer flipping frequency for a question
        \STATE Sample state $s_0\sim p_0(s_0)$
        \WHILE{$size(S) < M$}
            \STATE $d = $ environment description (describes the current scene using questions with T/F answers)
            \FOR{\textit{q} \textbf{in} \textit{d}}
            \IF{\textit{q} not in \textit{S}}
            \STATE add \textit{q} to \textit{S}
            \STATE $C[q] = 0$
            \ENDIF
            \ENDFOR
            \STATE reset environment

        \ENDWHILE

        \FOR{$i=1$ {\bfseries to} $K$}
            \FOR{$j=1$ {\bfseries to} $n$}
                \STATE sample $q \sim S$
                \STATE add $q$ to $D[i]$
                \STATE remove $q$ from \textit{S}
            \ENDFOR
        \ENDFOR
        \STATE $t=0$
        \FOR{$\beta=1$ {\bfseries to} $N$}
            \STATE intrinsic reward $r_t^i=0$
            \STATE shuffle entries in \textit{D}
            \FOR{$j=1$ {\bfseries to} $K$}
               \STATE $q_1, q_2, ... q_n = D[j]$
               \STATE evaluate $A(s_t, q_1), A(s_t, q_2), ... , A(s_t, q_n)$
               \STATE sample $a_t \sim \pi(a_t \mid s_t)$
               \STATE sample $s_{t+1}, r_t^e \sim p(s_{t+1}, r_t^e \mid s_t, a_t)$
               \STATE evaluate $A(s_{t+1}, q_1), A(s_{t+1}, q_2), ... , A(s_{t+1}, q_n)$
               \FOR{$k=1$ {\bfseries to} $n$}
               \IF{$A(s_t, q_k) \neq A(s_{t+1}, q_k)$}
               \STATE $r_t^i$ += 1
               \STATE $C[q_k]$ += 1
               \IF{$C[q_k]\,\,/\,\,\beta \geq \alpha$}
               \STATE replace $q_k$ with new question \textit{q} from \textit{S} (question at index 0)
               \STATE remove \textit{q} from \textit{S}
               \ENDIF
               \ENDIF
               \ENDFOR
               \STATE add $s_t, s_{t+1}, a_t, r_t^e, r_t^i$ to optimization batch $B_{\beta}$
               \STATE t += 1
            \ENDFOR
            \STATE Calculate target $T_{\beta}$ and advantage $Adv_{\beta}$ 
            \FOR{$j=1$ {\bfseries to} $N_{\text{opt}}$}
            \STATE optimize $\theta_{\pi}$ wrt PPO loss on batch $B_{\beta}, T_{\beta}, Adv_{\beta}$ using Adam
            \ENDFOR
        \ENDFOR
    \end{algorithmic}
    \end{algorithm}

\vspace{-2mm}
\section{Experimental Details}
\label{app_details}
\vspace{-2mm}

We use one copy of the environment since CLEVR-Robot Environment does not support multithreading currently. We used rollouts of length $128$ in all experiments. We use $3$ optimization epochs per rollout for our approach and ICM, whereas $4$ epochs for RND. The episode terminates either if the agent achieves the goal or exceeds maximum time steps. The agent is provided a sparse terminal binary reward only if it arranges the objects according to the spatial relationship defined by the goal (e.g., arrange the objects horizontally according to some color ordering), and 0 otherwise.

Table \ref{table:preprocessing_env} contains details of how we preprocessed the environment for our experiments.

\begin{table}[ht]
\centering
\begin{tabular}{c | c}
 Hyperparameter & Value  \\ [0.5ex]
 \hline
 Grey-scaling & False  \\
 Observation downsampling & (64,64)  \\
 Extrinsic reward clipping & False \\
 Intrinsic reward clipping & False 
 
\end{tabular}
\caption{Preprocessing details for the environments for all experiments.}
\label{table:preprocessing_env}
\end{table}

We refer to the following open-source repositories for baselines:

ICM: \href{https://github.com/pathak22/noreward-rl}{https://github.com/pathak22/noreward-rl}\\
RND: \href{https://github.com/openai/random-network-distillation}{https://github.com/openai/random-network-distillation}\\
ICM (Pytorch implementation): \href{https://github.com/jcwleo/curiosity-driven-exploration-pytorch}{https://github.com/jcwleo/curiosity-driven-exploration-pytorch}\\
RND (Pytorch implementation): \href{https://github.com/jcwleo/random-network-distillation-pytorch}{https://github.com/jcwleo/random-network-distillation-pytorch}




\vspace{-2mm}
\section{Additional Results}
\label{app_results}
\vspace{-2mm}

We present additional analysis on several design decisions in our approach.

\vspace{-2mm}
\subsection{Varying number of questions}
\label{app_multiple_ques}
\vspace{-2mm}

\begin{figure}[!ht]
\begin{minipage}{.49\textwidth}
\centering
\vspace{-2mm}
\includegraphics[width=\linewidth]{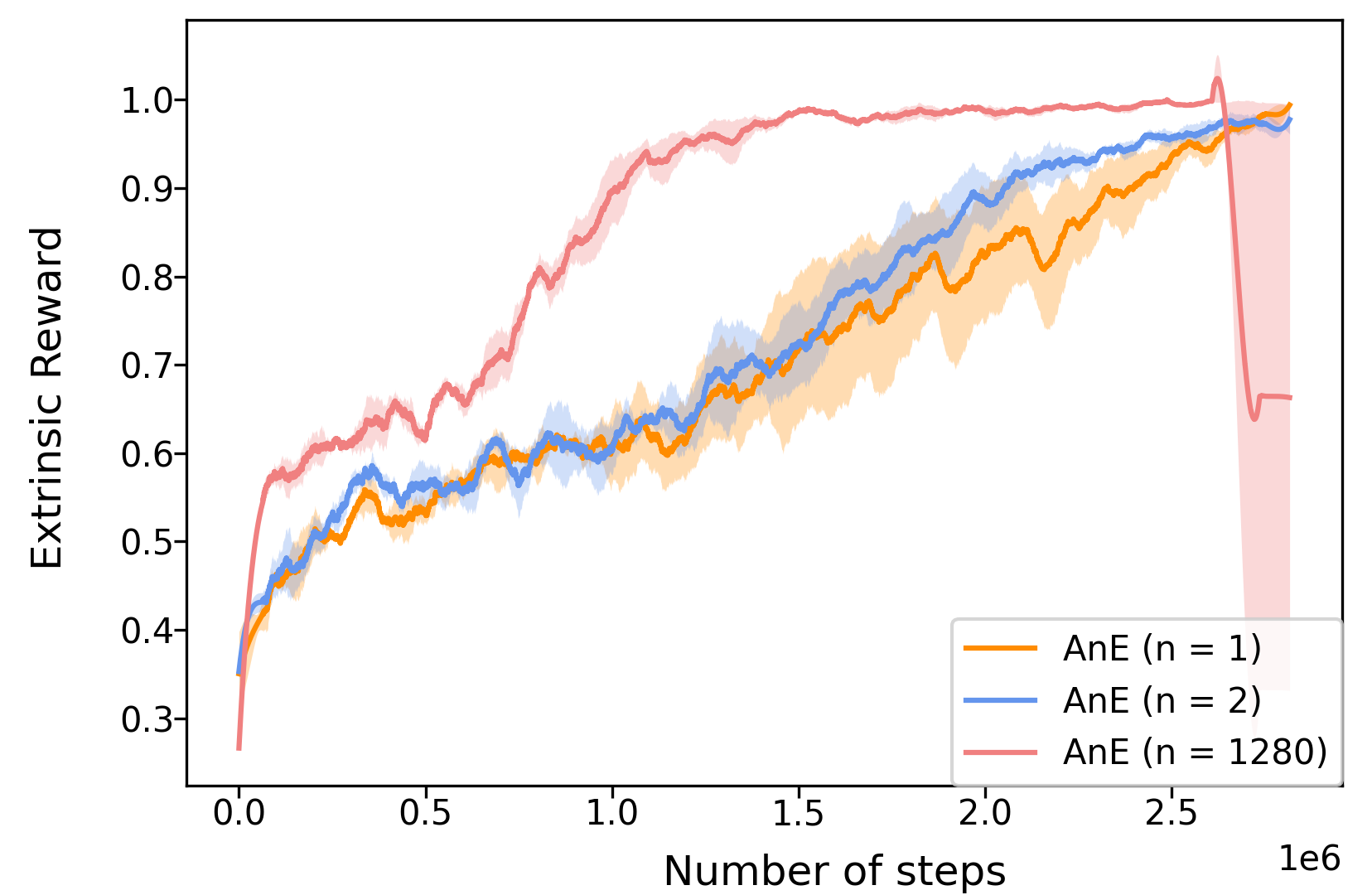}
(a) Dense reward setting
\end{minipage}%
\hspace{.02\textwidth}
\begin{minipage}{.49\textwidth}
\centering
\vspace{-2mm}
\includegraphics[width=\linewidth]{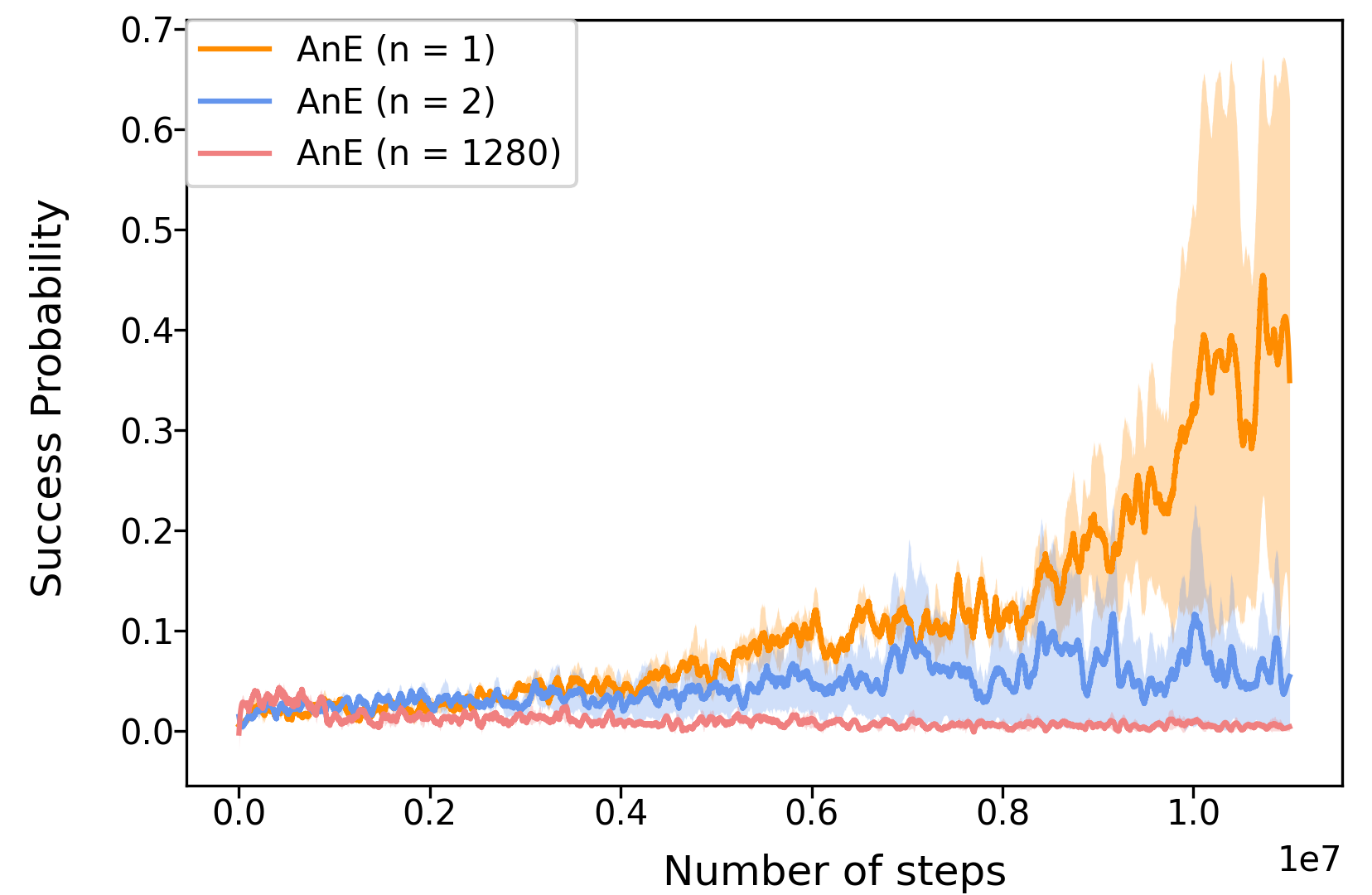}
(b) Sparse reward setting
\end{minipage}%
\caption{Comparing the performance of \names\ for different values of $n$. \vspace{-2mm}}
\label{fig:density_plots}
\end{figure}

We study the impact of number of questions used by the agent to query the environment on its performance in both dense and sparse reward settings. The results are shown in Figure~\ref{fig:density_plots}. 

It is interesting to observe that effect of linguistic feedback's density changes across different settings. For object alignment tasks in the dense reward setting, the agent's performance has an increasing relationship with the number of questions $(n)$ asked at each step. On the contrary, in the sparse reward settings, we notice that increasing the number of questions does not help the agent explore better, and the agent's curiosity is declining with increasing $n$. We believe this difference can be attributed to the highly different nature of the tasks in terms of both complexity and also the potential inherent impossibility of  simultaneously achieving high intrinsic reward and extrinsic reward -- this effect is magnified when the reward is extremely sparse (in the ordering task), so a trade-off is needed to be made in terms of linguistic feedback density. It would be interesting to see if it is possible to automatically find such balance.

\vspace{-2mm}
\subsection{VQA Model for curiosity-driven exploration}
\label{app_vqa}
\vspace{-2mm}

\begin{wrapfigure}{r}{0.45\textwidth}
\vspace{-6mm}
  \begin{center}
    \includegraphics[width=0.45\textwidth, clip]{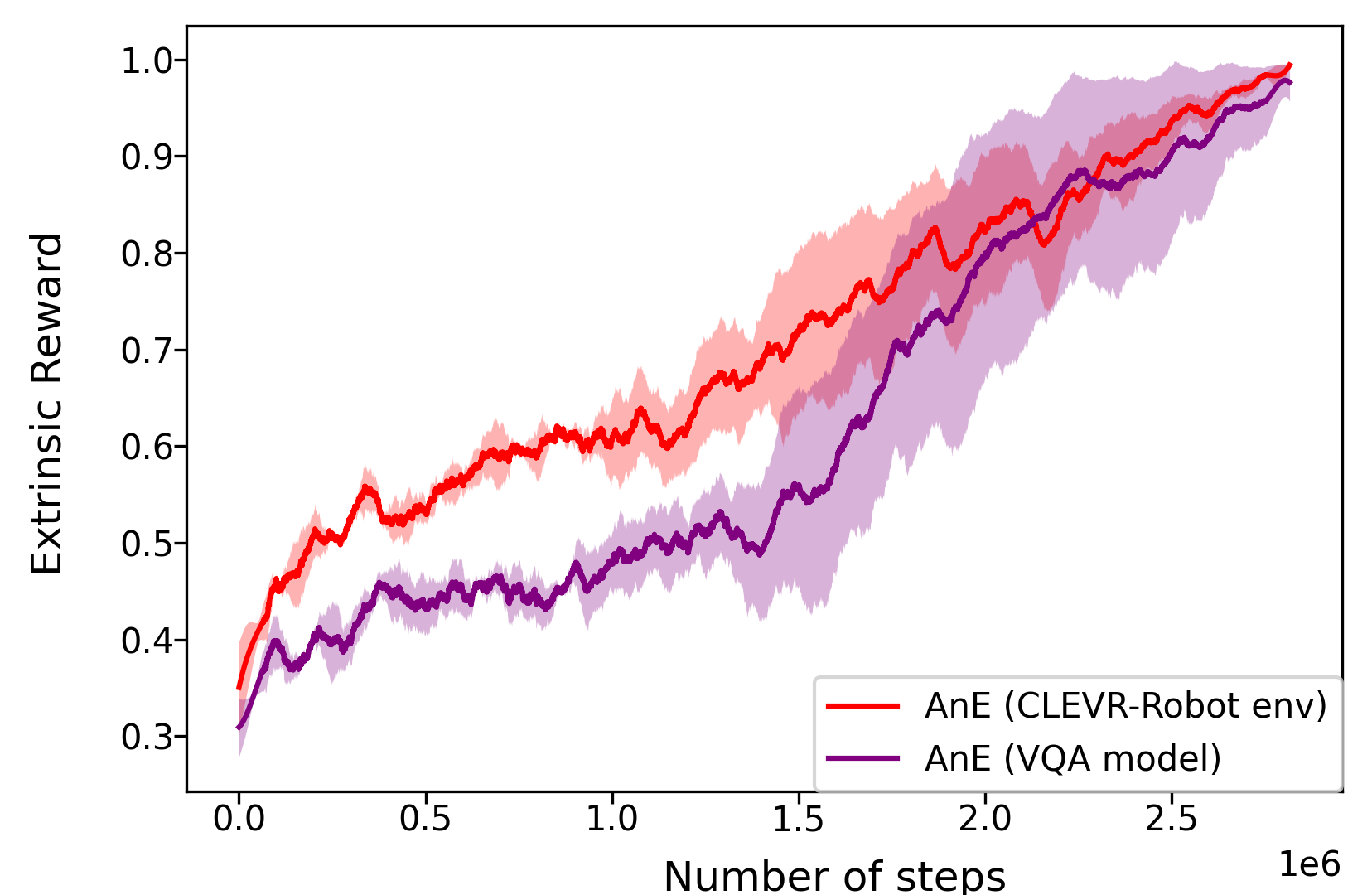}
  \end{center}
  \vspace{-0.4cm}
  \caption{Performance of \names\ using a VQA model for grounded question answering}
  \vspace{-6mm}
\label{fig:vqa_plot}
\end{wrapfigure}

We demonstrate that an agent equipped with a parameterized VQA model possesses grounded language understanding and hence can leverage our approach for curiosity-driven exploration (Figure \ref{fig:vqa_plot}). We train a CNN-LSTM model used as a baseline in~\citet{johnson2017inferring}.
We plan to work with more sophisticated and interpretable approaches in future works which represent human language as programs~\citep{johnson2017inferring, yi2018neural} to scale up to diverse object settings and eventually to settings where ground truth language is not available such as navigation.

\vspace{-2mm}
\subsection{Pure Exploration}
\label{app_intrinsic_only}
\vspace{-2mm}

\begin{wrapfigure}{r}{0.45\textwidth}
\vspace{-6mm}
  \begin{center}
    \includegraphics[width=0.45\textwidth, clip]{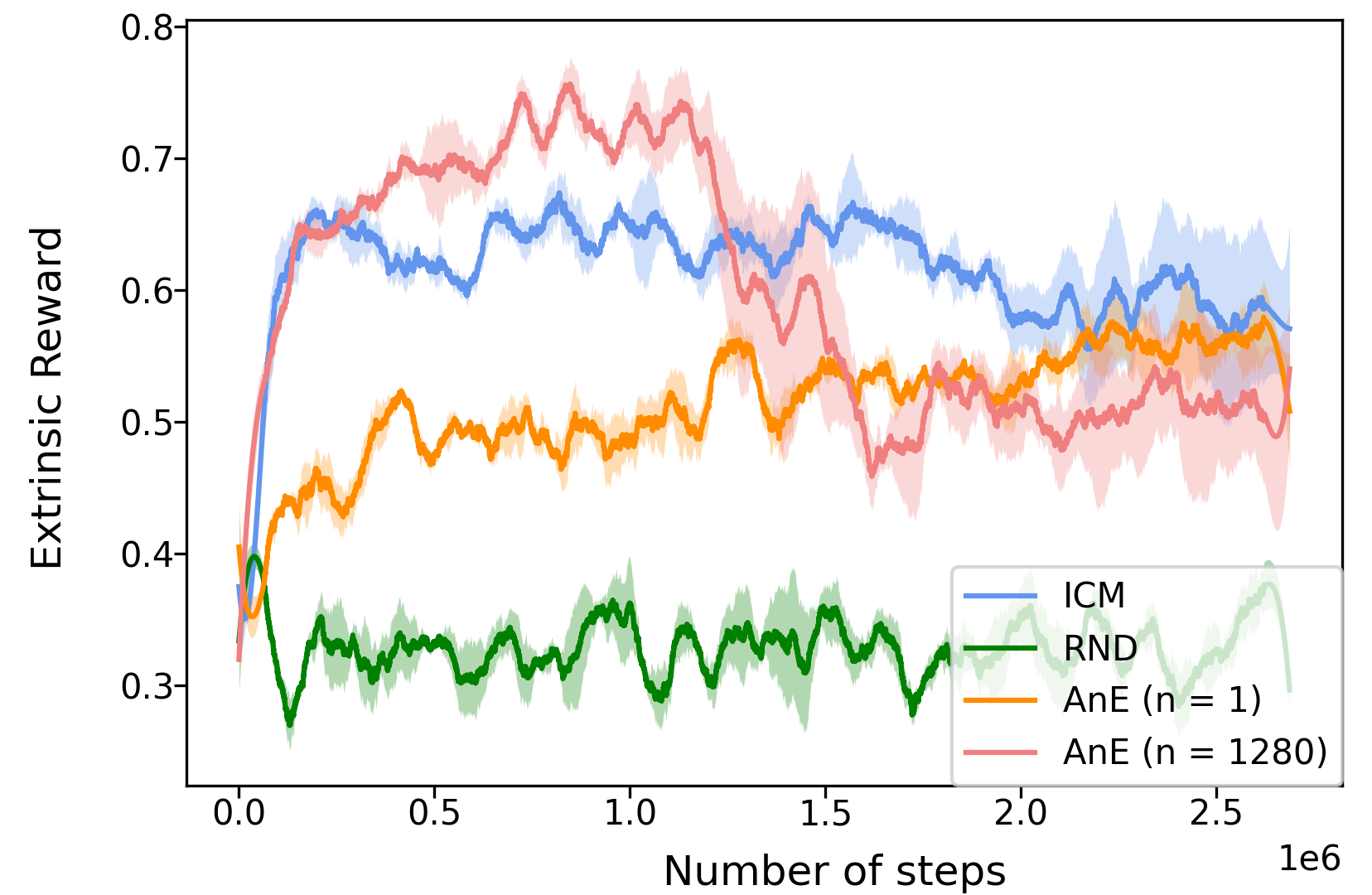}
  \end{center}
  \vspace{-0.4cm}
  \caption{Comparing the performance of \names\ against baselines ICM and RND in the absence of any extrinsic reward}
  \vspace{-6mm}
\label{fig:pure_exploration}
\end{wrapfigure}

We compare the performance of \names\ against curiosity-based baseline methods ICM and RND using pure exploration agents (agent does not have access to extrinsic reward) (Figure \ref{fig:pure_exploration}). We observe that even in the absence of extrinsic reward \names\ performs comparable to ICM and better than RND in the dense reward setting. In the sparse reward settings, all methods were unable to achieve success solely using the intrinsic reward, which suggests that the effect of \names\ comes from more than having denser extrinsic reward signal.

\end{document}

%% file: math_commands.tex
\usepackage{amsmath,amsfonts,bm}

\def\eqref#1{eq~(\ref{#1})}

\def\1{\bm{1}}

\DeclareMathAlphabet{\mathsfit}{\encodingdefault}{\sfdefault}{m}{sl}
\SetMathAlphabet{\mathsfit}{bold}{\encodingdefault}{\sfdefault}{bx}{n}

\usepackage{hyperref}

\usepackage{booktabs}       %
\usepackage{amsfonts}       %
\usepackage{nicefrac}       %
\usepackage{microtype}      %
\usepackage{subcaption}

\usepackage{enumitem}
\setlist{nolistsep}
\setlist[itemize]{noitemsep, topsep=0pt}

\usepackage{times}

\usepackage{graphicx} %
\usepackage{color}

\usepackage{array}
\usepackage{amssymb}
\usepackage{amsmath}
\usepackage{xspace}
\usepackage{fancyhdr}
\usepackage{comment}
\usepackage{bbm}

\newcolumntype{H}{>{\setbox0=\hbox\bgroup}c<{\egroup}@{}}

\newcommand{\noaistats}[1]{}  %

\definecolor{darkgreen}{rgb}{0,0.4,0.0}
\definecolor{darkblue}{rgb}{0,0.1,0.3}
\definecolor{darkred}{rgb}{0.7,0.0,0.0}

\renewcommand{\algorithmicensure}{}